\documentclass[sigconf]{acmart}

\usepackage{balance}

\usepackage{colortbl}
\usepackage[flushleft]{threeparttable}
\usepackage{adjustbox}
\usepackage{subcaption}
\usepackage[most]{tcolorbox}
\usepackage[utf8]{inputenc}

\def\BibTeX{{\rm B\kern-.05em{\sc i\kern-.025em b}\kern-.08emT\kern-.1667em\lower.7ex\hbox{E}\kern-.125emX}}
    
\copyrightyear{2019}
\acmYear{2019}
\setcopyright{acmcopyright}
\acmConference[SUMAC '19]{1st Workshop on Structuring and Understanding of Multimedia heritAge Contents}{October 21, 2019}{Nice, France}
\acmBooktitle{1st Workshop on Structuring and Understanding of Multimedia heritAge Contents (SUMAC '19), October 21, 2019, Nice, France}
\acmPrice{15.00}
\acmDOI{10.1145/3347317.3357246}
\acmISBN{978-1-4503-6910-7/19/10}

\begin{document}

\fancyhead{}

\title{Challenging Deep Image Descriptors for Retrieval in Heterogeneous Iconographic Collections}

\author{Dimitri Gominski}
\email{dimitri.gominski@ign.fr}
\affiliation{%
  \institution{Univ. Paris-Est, IGN-ENSG/LaSTIG}
  \streetaddress{73 Avenue de Paris}
  \city{Saint-Mandé, France}
  \postcode{94160}
}
\affiliation{%
  \institution{École Centrale Lyon - LIRIS}
  \streetaddress{36 Avenue Guy de Collongue}
  \city{Écully, France}
  \postcode{69134}
}

\author{Martyna Poreba}
\email{Martyna.Poreba@ign.fr}
\affiliation{%
  \institution{Univ. Paris-Est, IGN-ENSG/LaSTIG}
  \streetaddress{73 Avenue de Paris}
  \city{Saint-Mandé, France}
  \postcode{94160}
}

\author{Valérie Gouet-Brunet}
\email{Valerie.Gouet@ign.fr}
\affiliation{%
  \institution{Univ. Paris-Est, IGN-ENSG/LaSTIG}
  \streetaddress{73 Avenue de Paris}
  \city{Saint-Mandé, France}
  \postcode{F-94160}}

\author{Liming Chen}
\email{liming.chen@ec-lyon.fr]}
\affiliation{%
  \institution{École Centrale Lyon - LIRIS}
  \streetaddress{36 Avenue Guy de Collongue}
  \city{Écully, France}
  \postcode{69134}
}

%
\renewcommand{\shortauthors}{Trovato and Tobin, et al.}

%
\begin{abstract}
This article proposes to study the behavior of recent and efficient state-of-the-art deep-learning based image descriptors for content-based image retrieval, facing a panel of complex variations appearing in heterogeneous image datasets, in particular in cultural collections that may involve multi-source, multi-date and multi-view contents. For this purpose, we introduce a novel dataset, namely Alegoria dataset, consisting of 12,952 iconographic contents representing landscapes of the French territory, and encapsultating a large range of intra-class variations of appearance which were finely labelled. Six deep features (DELF, NetVLAD, GeM, MAC, RMAC, SPoC) and a hand-crafted local descriptor (ORB) are evaluated against these variations. Their performance are discussed, with the objective of providing the reader with research directions for improving image description techniques dedicated to complex heterogeneous datasets that are now increasingly present in topical applications targeting heritage valorization.
\end{abstract}

%
%


%

%
\begin{teaserfigure}
    \centering
    \includegraphics[scale=0.65]{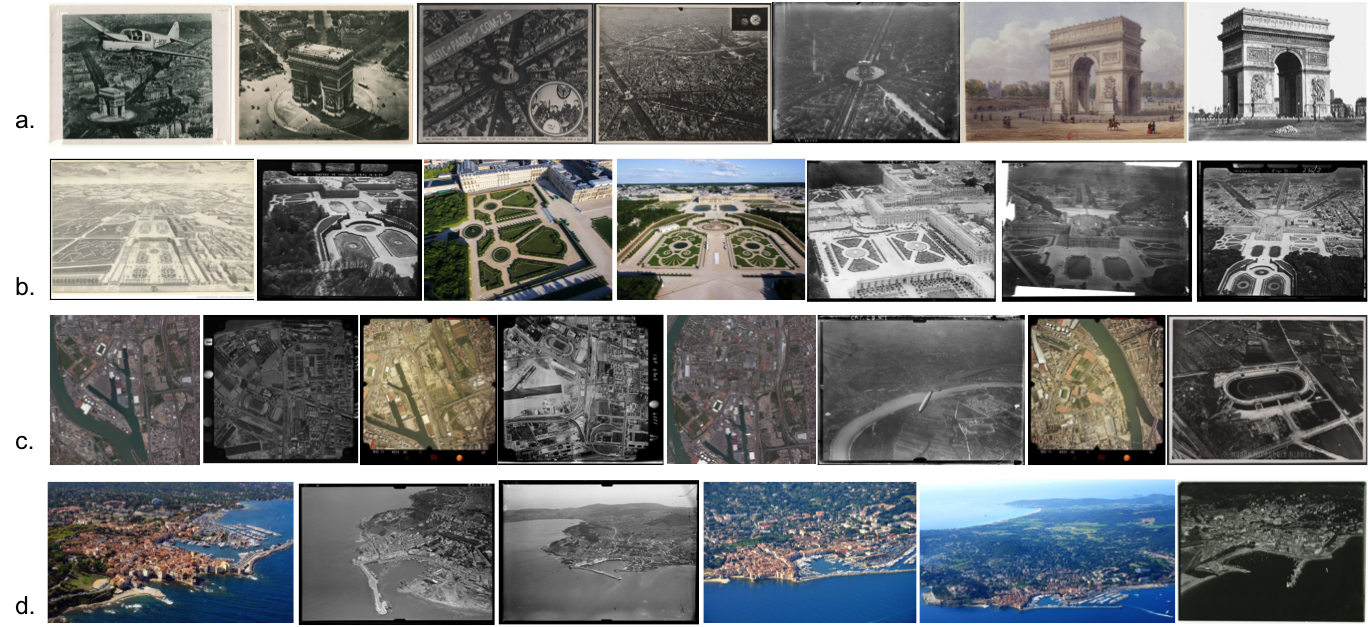}
    \caption{Examples from the Alegoria benchmark. Each row shows different images for the same scene landmark: a. Arc de Triomphe (class 20); b. Palace of Versailles (class 34); c. Gerland Stadium \& Port of Lyon (class 17); d. Dock of Saint-Tropez (class 39). \\
    \textit{Royalty-free images from gallica.bnf.fr - Bibliothèque nationale de France, and from IGN  (France) - Photothèque  Nationale. }}
    \label{fig:benchmark}
\end{teaserfigure}

%
\maketitle

\section{Introduction}
Digital or digitized cultural collections of images offer the interesting challenge of managing a wide variety of images together, mixing photographs with hand-made content, old black and white images with color pictures, low-quality scans or blurry photographs with digital pictures, etc. 
These collections are usually hosted within various institutions ({\it e.g.} GLAM - Galleries, Libraries, Archives and Museums), then organized in silos, without any interconnection between themselves while parts of them may address the same content ({\it e.g.} old paintings and recent photographs of Notre-Dame). The annotations associated may be of variable quality, or application-driven, making the interconnection of the fragmented contents not always easy. However, a better overall organization of these funds would be profitable in many areas where the complementarity of the available resources improves the analysis, ranging from SSH to landscape ecology, including urban planning and multimedia. In addition, there is currently a significant interest in the massive digitization of these heritage collections, with a desire to make them accessible to as many people as possible for multiple uses, associated with relevant structuring and consultation tools. 

In this context, content-based image retrieval (CBIR) offers a powerful tool for establishing connections between images, independently of their native organization in the collections.
Because of the variety of such contents in terms of acquisition source, viewpoint and illumination changes, evolution of the content across time, the key-word characterizing these collections is \textit{heterogeneity}. 
These constraints introduce difficulty in the determination of efficient and robust content-based descriptors, from pixel level (alterations due to the digitization process or the aging of photographic chemicals) to semantic level (is a place still considered the same place if every building has changed ?).

In this work, our objective is to establish an extensive evaluation of the most recent content-based descriptors, relying on deep features, for heterogeneous data such as those available in cultural collections, when considering an image retrieval task. This problem is sometimes referred to as "cross-domain" retrieval as in \cite{shrivastava_data-driven_2011} or \cite{bhowmik_cross-domain_2017}, but we will refrain from using that term since it suggests well-defined domains with their specific characteristics. Instead, we propose to speak of "heterogeneous" content-based image retrieval. 

The paper is organized as follows: in section \ref{se:hete_collections}, we revisit the characteristics of the main public image collections available and present the one we propose for experiments in this work, which is related to cultural heritage. Section \ref{se:deep_features} is dedicated to the presentation of recent deep learning based descriptors. They are experimented in section \ref{se:performance_evaluation}, where we discuss the impact of the photometric and geometric transformations available on the heterogeneous contents related to cultural collections, before concluding in the last section.

\section{Heterogeneous collections of images}\label{se:hete_collections}

This section is dedicated to image datasets involving very heterogeneous contents, with discussions about their main characteristics and the way of exploiting the latter in the qualitative and quantitative evaluation of image analysis and indexing techniques. We focus on cultural collections which gather interesting properties that continue to challenge the most efficient state-of-the-art image descriptors. 
In section \ref{se:Alegoria}, we present the Alegoria benchmark, which is a new challenging image dataset characterized by several interesting intra-class types of variations, before explaining in section \ref{se:annotation} how we model these variations in order to enable a sharp evaluation of state-of-the-art image descriptors. Section \ref{se:other_benchmarks} revisits other classical benchmarks and positions the Alegoria one alongside them.

\subsection{ALEGORIA dataset}\label{se:Alegoria}

To address the retrieval problem in heterogeneous collections, we propose a benchmark consisting of \textbf{12952} grey and color images of outdoor scenes. Each image (in JPEG format) has a maximum side length of 800 pixels. Street-view, oblique, vertical aerial images, sketches, or old postcards can be found, taken from different viewpoints, by different type of cameras at different periods and sometimes even under different weather conditions. These geographic iconographic contents describe the French territory at different times since the inter-war period to the present day. They contain multiple objects and cultural artifacts: buildings (also stadiums and train stations), churches and cathedrals, historical sites (\textit{e.g.} palaces, the most important monuments of Paris), seasides, suburbs of large cities, countrysides, \textit{etc.} Some example images are shown in Figure \ref{fig:benchmark}, each row represents the same geographical site.

To enable quantitative comparisons, a subset of \textbf{681} images of this database have been manually selected and labelled. There are \textbf{39} classes of at least 10 images, each one is associated with the same topics site, for example Eiffel Tower, Arc de Triomphe, Notre-Dame de Paris, Sacr\'e-Coeur Basilica, Palace of Versailles, Palace of Chantilly,  Nanterre, Saint-Tropez, Stadium Lyon Gerland, Perrache train station. This benchmark can be used for CBIR in several applications such as place recognition, image-based localization or semantic segmentation.

\subsection{Annotation of the appearance variations}\label{se:annotation}

The Alegoria dataset is a good illustration of a highly multi-source, multi-date and multi-view dataset. This heterogeneity allows to highlight significant variations of appearance, such as landscape transformations (site development, seasonal changes in vegetation), perspective (significant change in angle of view) or quality (color, B\&W or sepia old photos). In order to evaluate the impact of these variations on current approaches of image analysis and indexing, we annotated the Alegoria dataset by considering a set of the most common and important intra-class variations. A total of 10 variation types were taken into account, including the usual Scale, Illumination and Orientation changes, plus variations that are more specific to cultural heritage : Alterations (chemical degradations or damages on the photographic paper before digitization), Color domains (grayscale, sepia, etc.), Domain representation (picture, drawing, painting), Time changes (impact of large time spans) ; and general indicators of difficulty like Clutter, Positionning (when the main object of interest is not central to the picture) and Undistinctiveness (when the object of interest is not clear even to the human eye). Only variations presenting a high level of difficulty were counted: obviously there is always a degree of scale variation between two images of the same class, but we counted it only if it clearly adds difficulty when we visually compare two images. 

To quantify this \textit{variation predominance}, we use the following annotation process. For each class, we define a reference image, carefully chosen as the image depicting the object in the most common way in the class. For example, the second image of row a. on Figure \ref{fig:benchmark} is a good reference, because most of the images in this class are in grayscale, with orientations between horizontal and vertical, with overall low quality, etc. We then get the \textit{variation predominance} score by comparing all images in the class to the reference image, and measuring the frequency of occurrence: 0 when no image is concerned, 1 when anecdotal (one or less image from the class concerned), 2 when multiple occurrences are present, 3 when the variation is predominant (more than a third of the class). This also allows the computation of an overall difficulty score, that gives hints about the most difficult classes associated with multiple severe types of variations.

\subsection{Relations to other benchmarks}\label{se:other_benchmarks}

The standard benchmarks for evaluation of content-based image retrieval techniques dedicated to landscapes are historically Oxford5k \citep{philbin_object_2007}, Paris6k \citep{philbin_lost_2008}, and to a lesser extent Holidays \citep{nister_scalable_2006} and UKBench \citep{jegou_hamming_2008}. Recently, \citet{radenovic_revisiting_2018} proposed a revised version of Oxford and Paris datasets ($\mathcal{R}$Oxford and $\mathcal{R}$Paris), correcting mistakes in the annotation and proposing three protocols of evaluation with varying levels of difficulty. 
The main variations in these small datasets (55 queries in 11 classes) arise due to image capture conditions like different viewpoints, occlusions, weather and illumination. 
Google also proposed its own dataset, namely Google-Landmarks \cite{noh_large-scale_2017}, which contains around 5 million images of more than 200000 different landmarks. But this dataset is for now mainly used for training descriptors. 

By introducing the Alegoria dataset, we aim at proposing a complementary benchmark, designed for precise evaluation of robustness on a broader panel of appearance variations. These variations bring into play challenging conditions such as long-term acquisitions (multi-date contents) as well as multi-source contents (drawings, engraving, photographs, etc.) that are not widely represented in the other popular datasets and have the additional interest of bridging cultural heritage and geographical information domains. We also generalize the content to a larger panel of geographical landscapes, including urban contents and landmarks as well as more natural landscapes such as mountains and rivers. The cathedral Notre-Dame de Paris is a good example of this complementarity: this landmark can be found in both Alegoria and Paris datasets, the difference being in what is evaluated. On Paris dataset, we assess the absolute performance of the retrieval method, whereas on Alegoria we can assess how the method reacts to different types of variations, including variations due to multi-date and multi-source contents.

\section{Deep features}\label{se:deep_features}

Recently, convolutional neural networks (CNNs), aided with GPUs \citep{krizhevsky_imagenet_2012}, have been proven to be powerful feature extractors. Contrary to the hand-crafted methods where descriptors were carefully designed to maximize invariance and discriminability ({\it e.g.} SIFT, ORB, SURF...), deep learning offers a way of letting an optimization algorithm to determine how to get these characteristics. In the seminal work of \citet{babenko_neural_2014}, features were simply obtained at the output of the fully connected layers in early networks such as Alexnet. \citet{babenko_aggregating_2015} proved that aggregating features from the last convolutional layers produced better global descriptors. But these techniques lacked the core advantage of deep learning : optimizing the network directly for the retrieval task.

\citet{arandjelovic_netvlad:_2017} resolved this problem by proposing an end-to-end learning of deep features. Also declined in local \citep{yi_lift:_2016}, or patch-based \citep{xufeng_han_matchnet:_2015} versions, these works completed the toolbox of deep features that was now ready to replace hand-crafted methods.

The backbone of recent image retrieval methods using deep features relies on a CNN, applying a function \(f_i\) to the input image \(I\) (depending on what layer \(i\) is considered), and producing a tensor of activations \(T_i\) : \(T_i = f_i(I)\). Through the training process, we aim at optimizing \(f_i\) so that this 3D tensor, with width \(W\) and height \(H\) depending on the dimensions of the input image, and depth \(D\) depending on \(i\), contain discriminative information. However \(T_i\) is too big to be indexed and compared during the retrieval process, it is thus mandatory to design a method reducing the memory cost of describing the images. The following methods will be presented along this guideline of how \(T_i\) is handled, giving either a local or a global descriptor.

\subsection{Local methods}

Local methods use a set of carefully selected points on an image. This involves identifying points that maximize invariance, and describing small patches around these points to extract information. 
Early deep methods replaced parts of the historical hand-crafted pipeline by trainable pieces. \citet{verdie_tilde:_2015} designed a learnable keypoint detector maximizing repeatability under drastic illumination changes.
\citet{yi_lift:_2016} extended the architecture for full point detecting and describing. However these two methods aim at building a robust detector/descriptor, whereas we want to \textit{fit} our features to the data.

\begin{figure}
    \caption{Deep features extraction}
    \label{fig:deepfeat}
    \vspace{2mm}
\begin{subfigure}{\linewidth}
\centering
    \includegraphics[width=\textwidth]{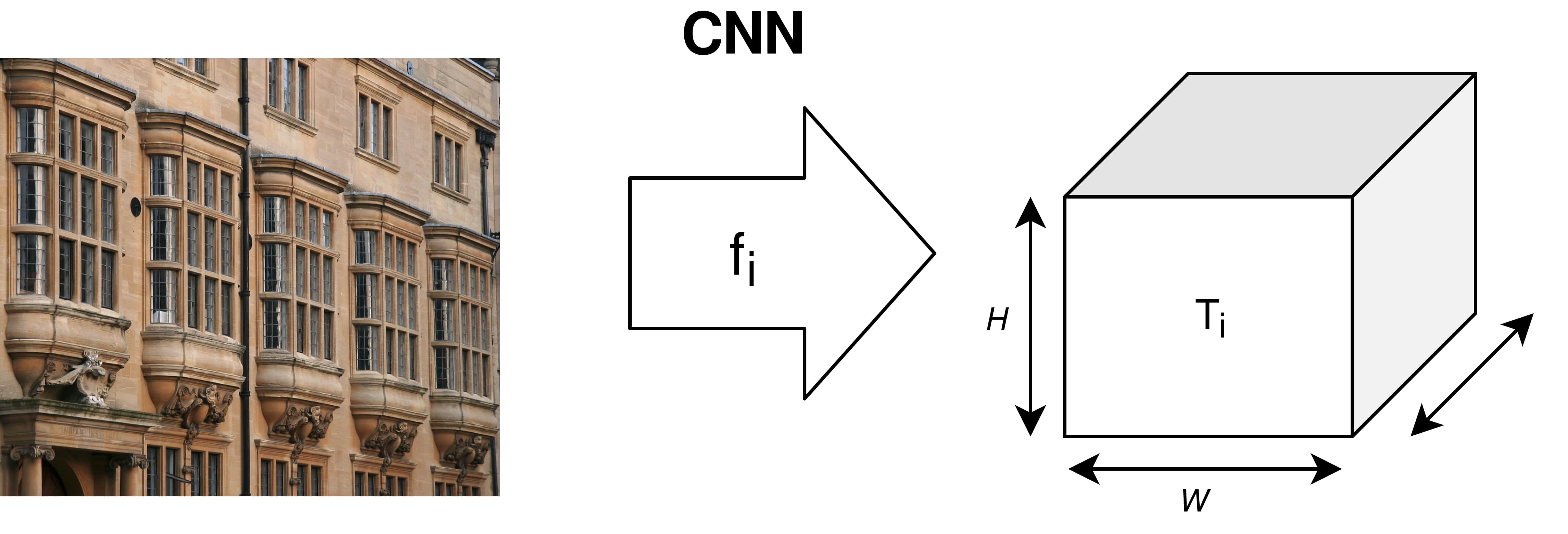}
\end{subfigure}\vspace{2mm}

  \begin{subfigure}{\linewidth}
    \centering
    \includegraphics[width=\textwidth]{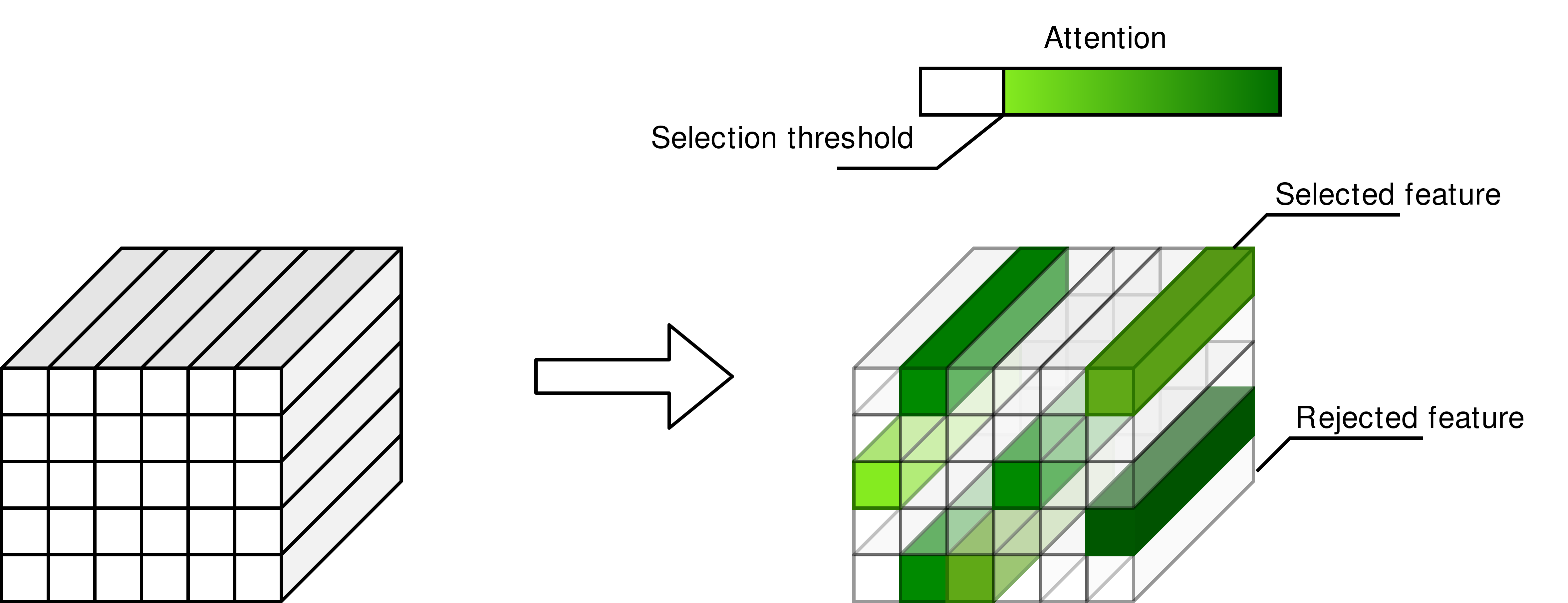}
    \caption{DELF feature extraction}
    \label{fig:delf}
\end{subfigure}\vspace{5mm}

  \begin{subfigure}{\linewidth}
    \centering
    \includegraphics[width=\textwidth]{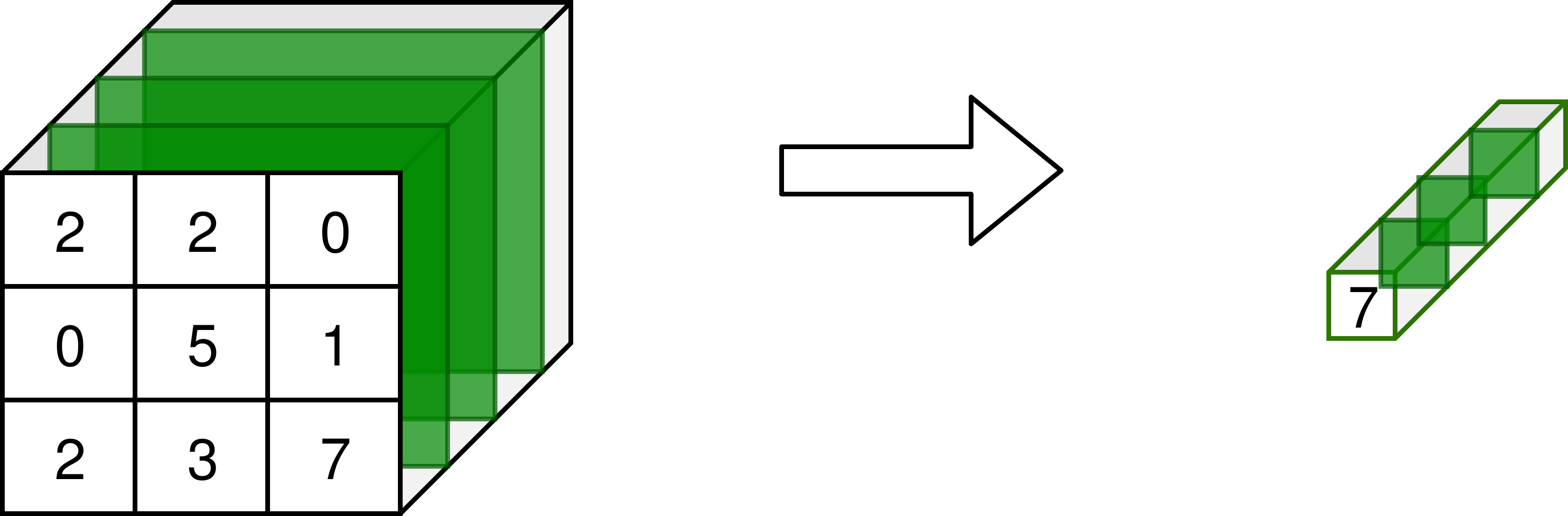}
    \caption{MAC descriptor}
    \label{fig:mac}
\end{subfigure}
\end{figure}

\citet{noh_large-scale_2017} solved the issue with a pipeline producing a set of local descriptors in a single forward pass, and that can be trained directly on any dataset with image-level labels. In their method, \(T_i\) is seen as a dense grid of local descriptors, where each position in the activation map is a \(D\)-dimensional local feature, whose receptive field in the input image is known. Additionally, an on-top function assigns a relevance score to each feature, and a threshold is set to only select most meaningful features. The output is a set of N \textbf{DELF} descriptors per image. See Figure \ref{fig:delf} for a visual interpretation. 
This departs from the traditional detect-then-describe process by selecting points \textit{after} describing them, but it is simple to train and has shown good results on standard benchmarks \citep{radenovic_revisiting_2018}.  Note that \(D\) typically ranges from 512 to 2048 in the last layers of the CNN, hence the PCA reduction to \(D=40\) proposed by the authors. Optimization on relevant data is done uniquely with a classification loss. 

\citet{dusmanu_d2-net:_2019} expanded this work with a detect-and-describe approach where they enforce keypoint repeatability and descriptor robustness using a Structure-from-Motion dataset with correponding points on different images. 

To perform the database similarity measure, local features must be aggregated. This is usually done with the learning of a dictionary (as in the well-known bag-of-words \citep{sivic_video_2003}), image are then described with a sparse vector suited for the inverted index structure. 

Some local methods aggregate local descriptors into a compact representation, like the VLAD descriptor \citep{jegou_aggregating_2010}.
\citet{arandjelovic_netvlad:_2017} mimic VLAD with a learnable pooling layer, giving \textbf{NetVLAD}. By replacing the hard assignment step with soft assignment to multiple clusters, they can train this layer. In this work, \(T_i\) is considered, as in \textbf{DELF}, as a block containing \(WxH\) D-dimensional descriptors.
The recent work of \citet{teichmann_detect--retrieve:_2018} builds upon the same idea, they describe selected candidate regions in the query with VLAD, and then propose a regional version of the ASMK \cite{tolias_image_2016} (another aggregation method) to aggregate these descriptors into a global descriptor.

Finally, the also recent work of \citet{simeoni_local_2019} proposes a new view, following the observation that shapes of objects of interest in the input image can be found in some channels of \(T_i\). They perform detection and description of interest points in \(T_i\) using a hand-crafted detector (MSER \citep{matas_robust_2004}), and then match images based on spatial verification. However, this method is not suitable for large-scale retrieval since it works on pairs of images.

\subsection{Global methods}

Global methods describe an image as a whole, embedding all important information in a single vector. This is conceptually closer to the classification task for which common architectures like Resnet \citep{he_deep_2016} or VGG \citep{simonyan_very_2014} were designed. \citet{babenko_neural_2014} indeed showed that simply taking intermediate features from a classification network and using them for retrieval yields good performance. 

To handle varying sizes in images and allegedly get more invariance, a standard seems to have emerged in deep global descriptors : extracting information at one of the last layers with a pooling process giving one value per channel. Following our notation, here the tensor \(T_i\) is seen as a set of D activation maps that contain each a different type of highly semantic information. To get a global descriptor, a straightforward approach is thus to get the most meaningful value per channel. We can then compare two images simply with dot-product similarity of their descriptors. See figure \ref{fig:mac} for an example with the MAC descriptor detailed below.
\citet{babenko_aggregating_2015} propose to sum the activations per channel, establishing the \textbf{SPoC} descriptor. This is equivalent to an average pooling operation. Differently, \citet{kalantidis_cross-dimensional_2016} tweak the SPoC descriptor with a spatial and channel weighting, while \citet{tolias_particular_2016} get better results by using the maximum value per channel (\textbf{MAC} descriptor). They also propose the regional version \textbf{RMAC}, by sampling windows at different scales and describing them separately.
\citet{radenovic_fine-tuning_2019} generalize the preceding approaches with a generalized mean pooling (\textbf{GeM}) including a learnable parameter.

Global methods allow efficient fine-tuning on relevant data, either with the triplet loss \citep{hoffer_deep_2015} or the contrastive loss \citep{chopra_learning_2005}.

\section{Performance evaluation}\label{se:performance_evaluation}

This section is dedicated to the evaluation of the most efficient and recent approaches of image description for image retrieval, revisited in section \ref{se:deep_features}, on the benchmark Alegoria presented in section \ref{se:hete_collections}, with the ambition of highlighting their behavior according to several types of appearance variations.

\begin{table*}
  \caption{Correlation between variations and performance}
  \label{tab:correlation}
  \begin{adjustbox}{width=0.65\textwidth}
\begin{tabular}{l|rrrrrrr}
\toprule
 & DELF & ORB & NetVLAD & GeM & MAC & RMAC & SPoC \\
\midrule
Scale                 & -0.41 & -0.36 & -0.47 & -0.45 & -0.47 & -0.50 & -0.43 \\
Illumination          & -0.32 & -0.39 & -0.53 & -0.48 & -0.45 & -0.49 & -0.45 \\
Orientation           & -0.42 & -0.37 & -0.50 & -0.46 & -0.44 & -0.49 & -0.45 \\
Color                 & -0.28 & -0.14 & -0.09 & -0.58 & -0.58 & -0.52 & -0.54 \\
Representation domain & 0.07  & -0.23 & -0.10 & -0.22 & -0.18 & -0.15 & -0.23 \\
Occlusion             & 0.05  & -0.26 & -0.38 & -0.33 & -0.37 & -0.34 & -0.32 \\
Positioning           & -0.17 & 0.07  & -0.13 & -0.38 & -0.41 & -0.41 & -0.33 \\
Clutter               & -0.13 & 0.02  & 0.11  & -0.10 & -0.06 & 0.00  & -0.10 \\
Undistinctiveness     & -0.02 & 0.09  & 0.09  & 0.22  & 0.16  & 0.24  & 0.23  \\
Alterations           & -0.31 & -0.07 & -0.15 & -0.23 & -0.22 & -0.19 & -0.23 \\
Time changes          & 0.05  & 0.23  & 0.35  & 0.19  & 0.13  & 0.13  & 0.22 \\
Overall difficulty    & -0.42 & -0.31 & -0.41 & -0.62 & -0.64 & -0.62 & -0.58
\end{tabular}
\end{adjustbox}
\end{table*}

\begin{table}
\caption{Color experiment: influence on the intra-class color variation on several descriptors.}
\label{ta:color}
\begin{tabular}{l|lll}
Global mAP & DELF & GeM & RMAC \\
\hline
Mixed color domains & 0.402 & 0.277 & 0.294 \\
Grayscale only & 0.421 & 0.282 & 0.299
\end{tabular}
\end{table}

We evaluate the performance of \textbf{DELF} and \textbf{NetVLAD} for deep local-based descriptors, and \textbf{GeM}, \textbf{MAC}, \textbf{RMAC} and \textbf{SPoC} for global-based descriptors. We also include the hand-crafted descriptor \textbf{ORB} for reference.

Since there is no dataset for fully training a deep feature network with heterogeneous data involving all the types of variations we consider, we evaluate methods in an off-the-shelf manner, with no fine-tuning. However, all these methods were trained on contents close to the Alegoria contents: DELF was trained on the Landmarks dataset, which is a large-scale noisy dataset of landmarks with some typical variations such as Scale, Orientation and Occlusion. NetVLAD, GeM, MAC, RMAC and SPoC were all trained using the code provided by \citet{radenovic_revisiting_2018}, on the retrieval SfM 120k dataset. This one also focused on specific objects, mostly landmarks, with also interior pictures and standard variations (mostly Scale, Orientation, Illumination). We use Resnet101 as the backbone architecture, giving 2048-dimensionnal descriptors, except for NetVLAD, for which we use the standard parameters of the original paper (64 clusters * 512 channels in \(T_i\)).

For fair comparison, we discard any post-processing step. Global methods are compared with simple dot-product. DELF (dimension 1024 as in the original paper without PCA) and ORB descriptors are matched one-to-one giving a number-of-inliers score assessing similarity, using a product-quantized index for efficient memory management.

To compare the image descriptors evaluated, we use the classical mAP score, computed per class. For each query \(q\), the average precision (AP) is computed on the sublist of results from 1 to \(k\), \(k\) being the index of the last positive result. 

The mAP per class is obtained by averaging the AP over all query images from a single class.

\subsection{Results}

The reader can refer to table \ref{tab:results} for the full lists of mAPs computed per class and the associated evaluation of variation predominance. To give an indicator of the overall difficulty of each class, we summed the predominance score of all the variations in the last column. 
We also computed the correlation matrix (see table \ref{tab:correlation}) between the results of each methods and the predominance scores, considering each column as a series of 39 observations. Lower values indicate negative correlation: this variation is highly correlated with a decrease in performance. This does not imply causality but gives insights on the correspondences between variations and the performance of the descriptor. Note that there are also positive values, notably for Time changes and Undistinctiveness, indicating that these variations are correlated with other factors that on the contrary improve performance. This might be a result of a bias in the selection of the pictures: when selecting pictures on a long time range for example, we tend to reduce the actual difficulty.

The Overall difficulty correlation score gives us a sanity check : overall difficulty indeed has a consistent negative correlation with the performance of all methods. 

\subsection{Discussion}

In this section, we discuss the results obtained according to several criteria:

\subsubsection{Local vs global description} The local DELF descriptor yields the best results with a consistent and significant margin. Table \ref{tab:correlation} shows that it is particularly stronger than global methods against Occlusion and Representation domain changes. This highlights the well-known advantage of local methods: by focusing on a set of local areas, they avoid the semantic noise captured by global methods. They also avoid the usual centering bias of global methods, as shown with the better Positioning score. However, on classes consisting of aerial images ({\it e.g.}class 7) RMAC gets better results. We believe this is due to the training dataset of DELF: it does not contain much aerial images, and DELF enforces a selection of important keypoints with its learned attention mechanism. DELF thus fails to find enough discriminative points on this type of data, whereas RMAC captures information on a large part of the image, allowing better results. Figure \ref{fig:DELFclass08} gives an example where DELF fails to find true correspondences between two images from class 8.

\subsubsection{Pooling} NetVLAD, GeM, MAC and SPoC only differ in the way they pool the tensor \(T\) to get a single global descriptor. We note that GeM gets overall better results, this can be explained by the fact that it generalizes MAC and SPoC with a tunable parameter, getting the best of both methods. NetVLAD performs consistently worse than others.

\subsubsection{Attention mechanisms}  
RMAC has overall better performance than other global descriptors. We confirm the observation from the original paper \citep{tolias_particular_2016} that the Region Proposal mechanism (which is basically an attention mechanism) of RMAC allows it to outperform its simpler version MAC, and we note that this is also true against other pooling methods.   

\begin{figure}[htbp]
\begin{subfigure}{\linewidth}
\centering
    \includegraphics[width=\textwidth]{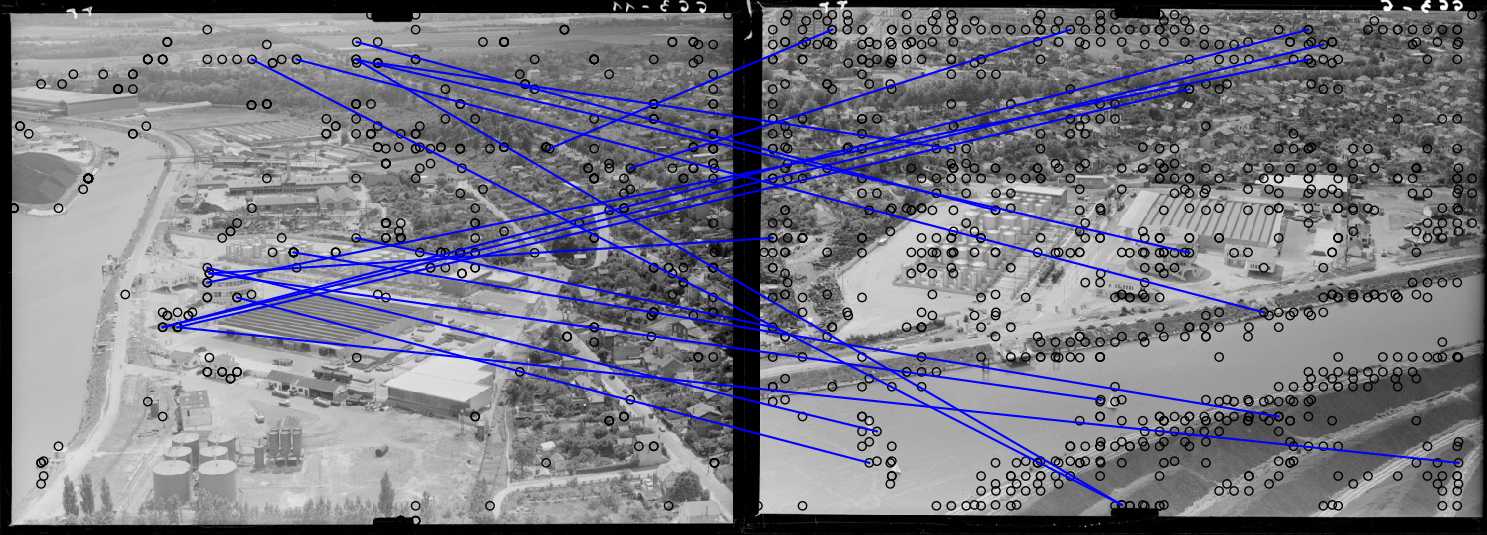}
    \caption{DELF matching failure case on aerial images}
    \label{fig:DELFclass08}
\end{subfigure}\vspace{3mm}

\begin{subfigure}{\linewidth}
\includegraphics[width=\textwidth]{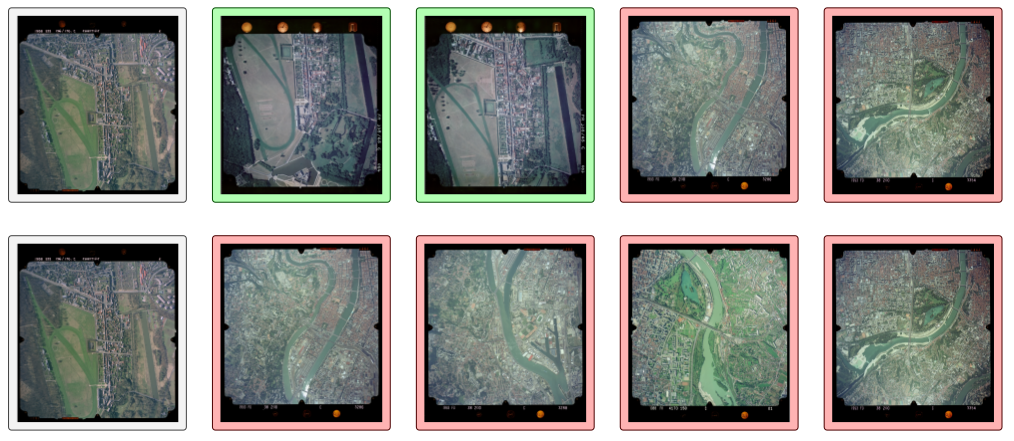}
\caption{RMAC vs. GeM descriptors on difficult cases. The top row shows the first 5 results for RMAC by decreasing order of similarity, the bottom row for GeM. The queries are in a white box, correct retrieved images in a green box and incorrect in a red box.}
\label{RMACGEM}
\end{subfigure}
\caption{Retrieval examples}
\end{figure}

See figure \ref{RMACGEM} for an example where RMAC gets better results because it focuses on the parts of the image considered to contain the object of interest, whereas GeM returns negative but visually similar images. 
As showed in the ablation studies in the original paper of DELF, its attention mechanism is also responsible for a performance boost, but we lack other deep local methods to highlight this fact on our data.

\subsubsection{Types of variations} Table \ref{tab:correlation} shows that Scale, Illumination, Orientation and Color are consistently associated with worse results. This shows that the main problem of image retrieval, even with modern deep learning methods, is still about getting invariance against basic variations. To support this analysis, we propose to do a simple experiment: mapping all the images in the same color domain (grayscale) before performing the description and the matching. We compute the global mAP for DELF, GeM and RMAC for the original dataset and for the grayscale dataset; see table \ref{ta:color} for the results. This normalization step reduces intra-class variance, but also the discriminative power of the descriptors. The mild but noticeable improvement shows that the former prevails, and we argue that a careful fine-tuning \citep{radenovic_fine-tuning_2019}
can maintain this discriminative power with reduced variance.  

\begin{table*}
  \caption{Results: retrieval accuracy against class variations}
  \label{tab:results}
  \begin{adjustbox}{width=1.0\textwidth}
  \begin{tabular}{lc|cccccccc|p{0.2cm}p{0.2cm}p{0.2cm}p{0.2cm}p{0.2cm}p{0.2cm}p{0.2cm}p{0.2cm}p{0.2cm}p{0.2cm}p{0.2cm}r}
  \toprule
    \multicolumn{2}{c}{} & \multicolumn{8}{|c|}{Descriptor} & \multicolumn{12}{c}{Variations} \\
    Class & \# & DELF & ORB & NetVLAD & GeM & MAC & RMAC & SPoC & Avg & \textit{Sc.} & \textit{Il.} & \textit{Or.} & \textit{Co.} & \textit{Do.} & \textit{Oc.} & \textit{Po.} & \textit{Cl.} & \textit{Un.} & \textit{Al.} & \textit{Ti.} & Overall \\
    \midrule
  1  & 48 & 0.249                         & 0.044 & 0.050 & \cellcolor[HTML]{32CB00}0.513 & 0.401                         & \cellcolor[HTML]{9AFF99}0.464 & \cellcolor[HTML]{34FF34}0.485 & 0.210 & 3 & 2 & 3 & 2 & 1 & 0 & 2 & 2 & 0 & 2 & 1 & 18 \\
2  & 13 & \cellcolor[HTML]{32CB00}0.807 & 0.091 & 0.095 & \cellcolor[HTML]{34FF34}0.370 & 0.306                         & \cellcolor[HTML]{9AFF99}0.357 & 0.296                         & 0.332 & 0 & 1 & 0 & 2 & 1 & 0 & 3 & 2 & 0 & 2 & 3 & 14 \\
3  & 16 & \cellcolor[HTML]{32CB00}0.604 & 0.171 & 0.071 & \cellcolor[HTML]{9AFF99}0.275 & 0.218                         & \cellcolor[HTML]{34FF34}0.303 & 0.263                         & 0.272 & 3 & 3 & 3 & 3 & 0 & 2 & 3 & 2 & 0 & 3 & 0 & 22 \\
4  & 22 & \cellcolor[HTML]{32CB00}0.389 & 0.054 & 0.050 & \cellcolor[HTML]{34FF34}0.231 & 0.186                         & 0.211                         & \cellcolor[HTML]{9AFF99}0.220 & 0.192 & 2 & 3 & 3 & 3 & 0 & 2 & 3 & 3 & 2 & 3 & 1 & 25 \\
5  & 18 & \cellcolor[HTML]{32CB00}0.270 & 0.059 & 0.060 & \cellcolor[HTML]{34FF34}0.236 & \cellcolor[HTML]{9AFF99}0.215 & 0.209                         & 0.211                         & 0.180 & 2 & 2 & 2 & 0 & 0 & 1 & 0 & 1 & 0 & 3 & 2 & 13 \\
6  & 11 & \cellcolor[HTML]{32CB00}0.623 & 0.269 & 0.210 & \cellcolor[HTML]{9AFF99}0.562 & 0.475                         & \cellcolor[HTML]{34FF34}0.588 & 0.538                         & 0.467 & 0 & 2 & 0 & 0 & 0 & 0 & 3 & 2 & 0 & 2 & 2 & 11 \\
7  & 11 & \cellcolor[HTML]{34FF34}0.729 & 0.123 & 0.099 & \cellcolor[HTML]{9AFF99}0.651 & 0.514                         & \cellcolor[HTML]{32CB00}0.822 & \cellcolor[HTML]{FFFFFF}0.626 & 0.509 & 0 & 0 & 0 & 0 & 0 & 0 & 0 & 3 & 3 & 3 & 0 & 9  \\
8  & 22 & 0.514                         & 0.082 & 0.075 & \cellcolor[HTML]{32CB00}0.661 & \cellcolor[HTML]{9AFF99}0.607 & \cellcolor[HTML]{34FF34}0.628 & 0.554                         & 0.446 & 0 & 2 & 0 & 0 & 0 & 0 & 0 & 3 & 0 & 3 & 3 & 11 \\
9  & 15 & \cellcolor[HTML]{32CB00}0.628 & 0.075 & 0.072 & \cellcolor[HTML]{9AFF99}0.333 & 0.235                         & 0.332                         & \cellcolor[HTML]{34FF34}0.338 & 0.288 & 3 & 3 & 2 & 2 & 0 & 2 & 3 & 1 & 0 & 1 & 0 & 17 \\
10 & 11 & 0.211                         & 0.093 & 0.093 & \cellcolor[HTML]{34FF34}0.233 & \cellcolor[HTML]{9AFF99}0.212 & \cellcolor[HTML]{32CB00}0.246 & 0.210                         & 0.185 & 3 & 2 & 3 & 3 & 0 & 0 & 3 & 3 & 3 & 3 & 3 & 26 \\
11 & 19 & \cellcolor[HTML]{32CB00}0.238 & 0.071 & 0.057 & \cellcolor[HTML]{9AFF99}0.153 & 0.138                         & \cellcolor[HTML]{34FF34}0.168 & 0.147                         & 0.139 & 2 & 3 & 0 & 3 & 0 & 0 & 3 & 3 & 3 & 3 & 0 & 20 \\
12 & 12 & \cellcolor[HTML]{34FF34}0.152 & 0.099 & 0.087 & \cellcolor[HTML]{32CB00}0.156 & 0.111                         & \cellcolor[HTML]{FFFFFF}0.150 & \cellcolor[HTML]{9AFF99}0.151 & 0.129 & 2 & 2 & 1 & 3 & 0 & 0 & 3 & 3 & 1 & 3 & 3 & 21 \\
13 & 13 & \cellcolor[HTML]{34FF34}0.296 & 0.136 & 0.099 & 0.273                         & 0.225                         & \cellcolor[HTML]{32CB00}0.338 & \cellcolor[HTML]{9AFF99}0.292 & 0.237 & 0 & 3 & 0 & 3 & 0 & 0 & 3 & 3 & 0 & 3 & 1 & 16 \\
14 & 10 & \cellcolor[HTML]{32CB00}0.519 & 0.142 & 0.128 & \cellcolor[HTML]{9AFF99}0.353 & 0.233                         & \cellcolor[HTML]{34FF34}0.403 & 0.348                         & 0.304 & 0 & 2 & 0 & 3 & 3 & 2 & 0 & 3 & 0 & 2 & 3 & 18 \\
15 & 14 & \cellcolor[HTML]{32CB00}0.149 & 0.079 & 0.076 & \cellcolor[HTML]{9AFF99}0.156 & \cellcolor[HTML]{34FF34}0.158 & 0.144                         & 0.133                         & 0.128 & 3 & 2 & 3 & 3 & 0 & 0 & 3 & 2 & 0 & 3 & 3 & 22 \\
16 & 26 & \cellcolor[HTML]{9AFF99}0.378 & 0.105 & 0.127 & \cellcolor[HTML]{34FF34}0.394 & 0.318                         & \cellcolor[HTML]{32CB00}0.412 & 0.327                         & 0.294 & 2 & 0 & 0 & 3 & 0 & 0 & 3 & 3 & 3 & 3 & 3 & 20 \\
17 & 11 & \cellcolor[HTML]{32CB00}0.157 & 0.092 & 0.097 & 0.115                         & \cellcolor[HTML]{34FF34}0.127 & 0.106                         & \cellcolor[HTML]{9AFF99}0.124 & 0.117 & 0 & 2 & 3 & 3 & 1 & 0 & 3 & 3 & 0 & 3 & 3 & 21 \\
18 & 19 & \cellcolor[HTML]{32CB00}0.467 & 0.209 & 0.121 & 0.335                         & 0.278                         & \cellcolor[HTML]{34FF34}0.353 & \cellcolor[HTML]{9AFF99}0.346 & 0.301 & 1 & 0 & 0 & 3 & 0 & 0 & 3 & 3 & 2 & 3 & 3 & 18 \\
19 & 27 & \cellcolor[HTML]{32CB00}0.167 & 0.044 & 0.042 & \cellcolor[HTML]{9AFF99}0.103 & \cellcolor[HTML]{34FF34}0.105 & \cellcolor[HTML]{34FF34}0.105 & 0.085                         & 0.093 & 3 & 3 & 3 & 3 & 2 & 0 & 2 & 3 & 0 & 3 & 0 & 22 \\
20 & 15 & \cellcolor[HTML]{32CB00}0.658 & 0.076 & 0.074 & 0.203                         & 0.160                         & \cellcolor[HTML]{34FF34}0.268 & \cellcolor[HTML]{9AFF99}0.235 & 0.239 & 1 & 3 & 0 & 3 & 2 & 2 & 2 & 3 & 0 & 2 & 3 & 21 \\
21 & 14 & \cellcolor[HTML]{32CB00}0.198 & 0.073 & 0.075 & 0.095                         & \cellcolor[HTML]{9AFF99}0.100 & \cellcolor[HTML]{34FF34}0.103 & 0.089                         & 0.105 & 2 & 3 & 3 & 3 & 2 & 2 & 3 & 3 & 0 & 3 & 3 & 27 \\
22 & 23 & \cellcolor[HTML]{32CB00}0.347 & 0.048 & 0.061 & \cellcolor[HTML]{9AFF99}0.110 & 0.104                         & \cellcolor[HTML]{34FF34}0.113 & 0.094                         & 0.125 & 3 & 3 & 3 & 3 & 2 & 2 & 2 & 3 & 0 & 3 & 0 & 24 \\
23 & 17 & \cellcolor[HTML]{32CB00}0.169 & 0.081 & 0.068 & 0.134                         & \cellcolor[HTML]{9AFF99}0.139 & \cellcolor[HTML]{34FF34}0.141 & 0.109                         & 0.120 & 3 & 2 & 1 & 3 & 2 & 2 & 2 & 3 & 0 & 3 & 0 & 21 \\
24 & 11 & \cellcolor[HTML]{32CB00}0.186 & 0.093 & 0.095 & \cellcolor[HTML]{9AFF99}0.174 & \cellcolor[HTML]{34FF34}0.181 & \cellcolor[HTML]{34FF34}0.181 & 0.146                         & 0.151 & 2 & 2 & 0 & 3 & 1 & 0 & 1 & 3 & 1 & 3 & 0 & 16 \\
25 & 12 & \cellcolor[HTML]{32CB00}0.560 & 0.086 & 0.126 & 0.449                         & \cellcolor[HTML]{9AFF99}0.452 & \cellcolor[HTML]{34FF34}0.541 & 0.382                         & 0.371 & 0 & 2 & 0 & 3 & 3 & 0 & 0 & 3 & 0 & 1 & 0 & 12 \\
26 & 26 & \cellcolor[HTML]{32CB00}0.322 & 0.047 & 0.046 & 0.111                         & \cellcolor[HTML]{34FF34}0.129 & \cellcolor[HTML]{9AFF99}0.114 & 0.067                         & 0.119 & 2 & 3 & 2 & 3 & 1 & 3 & 2 & 3 & 0 & 3 & 0 & 22 \\
27 & 36 & \cellcolor[HTML]{32CB00}0.268 & 0.037 & 0.041 & \cellcolor[HTML]{9AFF99}0.073 & 0.065                         & \cellcolor[HTML]{34FF34}0.089 & 0.060                         & 0.090 & 1 & 2 & 0 & 3 & 2 & 2 & 3 & 3 & 0 & 3 & 2 & 21 \\
28 & 14 & \cellcolor[HTML]{32CB00}0.507 & 0.105 & 0.099 & 0.239                         & \cellcolor[HTML]{9AFF99}0.247 & \cellcolor[HTML]{34FF34}0.259 & 0.200                         & 0.237 & 0 & 1 & 0 & 3 & 0 & 1 & 1 & 3 & 0 & 3 & 0 & 12 \\
29 & 21 & \cellcolor[HTML]{32CB00}0.297 & 0.058 & 0.066 & \cellcolor[HTML]{34FF34}0.254 & 0.192                         & 0.195                         & \cellcolor[HTML]{9AFF99}0.241 & 0.186 & 3 & 2 & 1 & 1 & 1 & 3 & 2 & 0 & 3 & 1 & 2 & 19 \\
30 & 11 & \cellcolor[HTML]{9AFF99}0.417 & 0.131 & 0.160 & \cellcolor[HTML]{34FF34}0.472 & 0.329                         & \cellcolor[HTML]{32CB00}0.566 & 0.450                         & 0.361 & 0 & 0 & 0 & 2 & 0 & 0 & 0 & 3 & 2 & 3 & 2 & 12 \\
31 & 16 & \cellcolor[HTML]{32CB00}0.353 & 0.064 & 0.074 & \cellcolor[HTML]{9AFF99}0.181 & 0.174                         & \cellcolor[HTML]{34FF34}0.184 & 0.153                         & 0.169 & 1 & 2 & 2 & 3 & 0 & 1 & 1 & 3 & 0 & 3 & 0 & 16 \\
32 & 30 & \cellcolor[HTML]{32CB00}0.197 & 0.105 & 0.107 & \cellcolor[HTML]{9AFF99}0.176 & 0.160                         & \cellcolor[HTML]{34FF34}0.193 & 0.135                         & 0.111 & 0 & 3 & 0 & 1 & 0 & 0 & 1 & 3 & 0 & 3 & 0 & 11 \\
33 & 17 & \cellcolor[HTML]{32CB00}0.715 & 0.111 & 0.117 & \cellcolor[HTML]{9AFF99}0.305 & 0.264                         & \cellcolor[HTML]{34FF34}0.362 & 0.292                         & 0.309 & 1 & 0 & 0 & 3 & 2 & 0 & 2 & 3 & 0 & 3 & 3 & 17 \\
34 & 10 & \cellcolor[HTML]{32CB00}0.284 & 0.121 & 0.104 & \cellcolor[HTML]{34FF34}0.240 & 0.216                         & \cellcolor[HTML]{9AFF99}0.231 & 0.213                         & 0.201 & 1 & 1 & 1 & 3 & 0 & 0 & 1 & 3 & 0 & 1 & 3 & 14 \\
35 & 10 & \cellcolor[HTML]{32CB00}0.317 & 0.102 & 0.103 & \cellcolor[HTML]{34FF34}0.193 & 0.176                         & \cellcolor[HTML]{9AFF99}0.186 & 0.170                         & 0.178 & 1 & 1 & 0 & 3 & 0 & 1 & 1 & 3 & 0 & 3 & 0 & 13 \\
36 & 18 & \cellcolor[HTML]{32CB00}0.309 & 0.070 & 0.065 & \cellcolor[HTML]{34FF34}0.197 & 0.158                         & \cellcolor[HTML]{9AFF99}0.194 & 0.186                         & 0.169 & 2 & 2 & 2 & 2 & 0 & 0 & 3 & 3 & 0 & 3 & 0 & 17 \\
37 & 18 & \cellcolor[HTML]{9AFF99}0.226 & 0.063 & 0.078 & \cellcolor[HTML]{32CB00}0.237 & 0.171                         & 0.225                         & \cellcolor[HTML]{34FF34}0.233 & 0.176 & 1 & 2 & 2 & 3 & 0 & 0 & 2 & 3 & 0 & 3 & 3 & 19 \\
38 & 13 & \cellcolor[HTML]{34FF34}0.446 & 0.110 & 0.155 & \cellcolor[HTML]{9AFF99}0.439 & 0.309                         & \cellcolor[HTML]{9AFF99}0.439 & \cellcolor[HTML]{32CB00}0.463 & 0.337 & 2 & 2 & 0 & 3 & 0 & 0 & 2 & 3 & 0 & 3 & 3 & 18 \\
39 & 11 & \cellcolor[HTML]{32CB00}0.294 & 0.095 & 0.097 & \cellcolor[HTML]{9AFF99}0.209 & 0.201                         & \cellcolor[HTML]{34FF34}0.216 & 0.200                         & 0.187 & 2 & 2 & 1 & 3 & 1 & 0 & 3 & 3 & 0 & 2 & 1 & 18
  \end{tabular}
  \end{adjustbox}
  \begin{tablenotes}
      \item \textbf{Best seen in color}
      \item \textit{Sc.}: Scale, \textit{Il.}: Illumination,  \textit{Or.}: Orientation, \textit{Co.}: Color, \textit{Do.}: Domain representation, \textit{Oc.}: Occlusion, \textit{Po.}: Positioning, \textit{Cl.}: Clutter, \textit{Un.}: Undistinctiveness, \textit{Al.}: Alterations, \textit{Ti.}: Time changes, \textit{Overall}: Overall difficulty. For each class, the 3 top mAPs are in green, ranging from dark (best result) to light green (3rd result). Column \textit{Avg.} indicates the average performance of all methods. Column \textit{Overall} gives an overall score of difficulty (sum of the variation scores).
    \end{tablenotes}
\end{table*}

\section{Conclusion}

We proposed a new benchmark for the evaluation of deep features on heterogeneous data, and discussed how most recent and efficient features react to the panel of variations encountered. Our results show that there is still many difficult cases to be handled by image retrieval methods, we thus presented insights on how to gain robustness with attention mechanisms and intra-class variance reduction. We believe this evaluation is necessary to allow image retrieval knowledge to be applied to real-world situations, and encourage future research to include detailed robustness studies, and to carefully design deep learning architectures for robust feature extraction regarding these variations.

\begin{acks}

This work is supported by ANR (French National Research Agency) and DGA (French Directorate General of Armaments) within the ALEGORIA project, under respective Grant ANR-17-CE38-0014-01 and DGA Agreement 2017-60-0048.

\end{acks}
%
\bibliographystyle{ACM-Reference-Format}
\bibliography{lib.bib}

%

\end{document}